\documentclass[12pt]{article}
\usepackage{graphicx}
\usepackage{amsmath}
\usepackage{amssymb}
\usepackage{physics}
\usepackage{authblk}
\usepackage{amsthm}
\usepackage{booktabs}
\usepackage{algorithm}
\usepackage{algpseudocode}

\usepackage{geometry}
\usepackage{setspace}
\usepackage{titlesec}
\usepackage[numbers]{natbib}  
\usepackage[colorlinks=true, linkcolor=black, citecolor=blue]{hyperref}

\titleformat{\section}[block]{\normalfont\Large\bfseries}{\thesection}{1em}{}
\titleformat{\subsection}[block]{\normalfont\large\bfseries}{\thesubsection}{1em}{}
\titleformat{\subsubsection}[block]{\normalfont\normalsize\bfseries}{\thesubsubsection}{1em}{}

\title{\textbf{Rethinking Over-Smoothing in Graph Neural Networks: \\ A Perspective from Anderson Localization}}

\author[1]{\small Kaichen Ouyang\thanks{Corresponding author: \texttt{oykc@mail.ustc.edu.cn}}}
\affil[1]{\small Department of Mathematics, University of Science and Technology of China, Hefei 230026, China}
\date{}

\begin{document}
\maketitle
\vspace{-1em}  

\begin{abstract}
Graph Neural Networks (GNNs) have shown great potential in graph data analysis due to their powerful representation capabilities. However, as the network depth increases, the issue of over-smoothing becomes more severe, causing node representations to lose their distinctiveness.This paper analyzes the mechanism of over-smoothing through the analogy to Anderson localization and introduces participation degree as a metric to quantify this phenomenon. Specifically, as the depth of the GNN increases, node features homogenize after multiple layers of message passing, leading to a loss of distinctiveness, similar to the behavior of vibration modes in disordered systems. In this context, over-smoothing in GNNs can be understood as the expansion of low-frequency modes (increased participation degree) and the localization of high-frequency modes (decreased participation degree).Based on this, we systematically reviewed the potential connection between the Anderson localization behavior in disordered systems and the over-smoothing behavior in Graph Neural Networks.A theoretical analysis was conducted, and we proposed the potential of alleviating over-smoothing by reducing the disorder in information propagation.
\end{abstract}


\section{Introduction}

Disordered systems are a central object of study in statistical physics, characterized by randomness or heterogeneity in their structural or dynamical properties. Classic examples include spin glasses\cite{binder1986spin}, amorphous solids\cite{zallen2008physics} and percolation networks\cite{li2021percolation}. These systems often exhibit rich collective phenomena, such as frustration, localization, and phase transitions, that are analytically challenging but crucial to understanding their macroscopic behavior.

Among these, a particularly significant phenomenon is \textbf{Anderson localization}\cite{anderson1958absence}, which describes how wavefunctions of particles, such as electrons, can become spatially localized due to random potentials in a disordered medium, leading to insulating behavior. The key feature of Anderson localization is the transition from a regime where particles or waves are free to propagate across the system (\textbf{metallic phase}) to one where wavefunctions are confined to localized regions, preventing effective transport (\textbf{insulating phase}). This transition can be quantitatively characterized by the concept of \textbf{participation degree}\cite{liu2010jamming}, which measures the spatial extent to which a mode or wavefunction spreads across the system.Interestingly, recent advances in machine learning suggest that modern neural networks—particularly deep and overparameterized architectures—can be viewed as a special class of disordered systems\cite{hopfield1982neural,sompolinsky1988chaos,baity2018comparing,amit1985spin,pennington2017geometry}. The complexity and randomness in weight initialization, data structure, and feature propagation bear strong analogies to the stochastic properties of physical disorder.

Neural networks, especially deep architectures, are increasingly studied through the lens of statistical physics\cite{bahri2020statistical,weng2023statistical,lee2024dynamic,kaplan2020scaling,cui2024phase}. Recent works have applied tools such as phase transition theory, spin-glass models, and replica methods to analyze complex learning behaviors.For instance, the phenomenon of \textit{grokking}\cite{rubin2023grokking} has been modeled as a first-order phase transition, capturing the sudden shift from memorization to generalization during training. Similarly, flow- and diffusion-based generative models exhibit sampling limitations analogous to phase boundaries in spin-glass systems\cite{ghio2024sampling}. In graph-based architectures such as Graph Neural Networks (GNNs), depth-related phenomena like over-smoothing have been analyzed using dynamical mean-field theory, revealing connections to localization\cite{duranthon2025statistical}. These insights suggest a deep structural analogy between neural computation and physical disorder, offering powerful frameworks for understanding learning dynamics and model behavior.

\textbf{Graph Neural Networks (GNNs)} represent a particularly successful and rapidly growing class of neural networks that operate on graph-structured data. GNNs have found applications in various domains, including recommendation systems\cite{fan2019graph}, biology\cite{jing2020learning}, and material\cite{reiser2022graph}, due to their ability to model relationships between entities and capture structural dependencies in graphs. However, despite their success, a key challenge that remains largely unresolved in GNNs is the phenomenon of \textbf{over-smoothing}, which occurs when the node features become homogenized across layers, causing the network to lose the ability to distinguish between different nodes. This issue arises because, in deeper networks, the aggregation process causes information from neighboring nodes to mix, gradually blurring the distinctions between node features.

Over-smoothing has significant implications for the performance of GNNs in tasks such as node classification, where the ability to preserve individual node features is crucial for differentiating between nodes\cite{oono2019graph}. Therefore, understanding the underlying causes of over-smoothing is essential for improving GNNs and mitigating its effects. While various empirical strategies have been proposed to address over-smoothing, a comprehensive theoretical understanding of the phenomenon is still lacking.

In this paper, we draw an analogy between over-smoothing in GNNs and \textbf{Anderson localization} in disordered systems, offering a new perspective on the phenomenon. Specifically, we borrow the concept of \textbf{participation degree} from the vibrational mode analysis of disordered systems, where it is used to quantify the spatial localization of eigenmodes. We adapt this metric to the context of GNNs to measure the extent of over-smoothing, drawing an analogy between the homogenization of node features and the localization behavior observed in physical systems.The contributions of this paper are as follows:
\begin{itemize}
    \item We reinterpret the over-smoothing phenomenon in GNNs as a localization effect and propose the \textbf{participation degree} as a principled metric for quantifying and understanding this phenomenon.
    \item We establish a structural and dynamic \textbf{duality} between disordered physical systems and graph neural networks, drawing on analogies to Anderson localization.
    \item We provide a theoretical framework that explains over-smoothing from a physics-inspired perspective, while also offering insights for architectural or training strategies to alleviate the issue.
\end{itemize}

\section{Related Work}
\subsection{Over-Smoothing in GNNs}

The over-smoothing phenomenon in Graph Neural Networks (GNNs) has emerged as a major challenge in the design and training of deep graph-based models. As the depth of a GNN increases, the node features tend to homogenize, resulting in a situation where the network struggles to differentiate between nodes.This issue has been observed in various studies, particularly when analyzing the behavior of GNNs as the number of layers grows~\cite{oono2019graph}.One of the key factors contributing to over-smoothing is the diffusion-like process of feature aggregation governed by the graph structure, typically captured by the graph Laplacian~\cite{chen2020measuring}.The Laplacian, which encodes information about the graph’s connectivity, drives the information diffusion process.When the network is deep, each layer aggregates features from an increasingly wide neighborhood, gradually blending the node features. This homogenization process erodes the unique characteristics of individual nodes and ultimately diminishes the network’s capacity for discriminative tasks such as node classification.

As GNNs become deeper, the influence of distant nodes accumulates layer by layer, causing the feature space to contract. In practice, this results in feature representations that are nearly indistinguishable across the graph, even for structurally or semantically distinct nodes. The effectiveness of message passing diminishes as the model converges toward a state where all nodes share similar embeddings, severely limiting the expressivity and learning capacity of the network.

Several approaches have been proposed to mitigate over-smoothing in GNNs:
\begin{itemize}
    \item \textbf{Residual Connections}\cite{li2019deepgcns}: Introducing residual or skip connections allows node features to be directly passed across layers without full aggregation, preserving important identity information and mitigating homogenization.
    \item \textbf{Edge Dropping}\cite{rong2019dropedge}: Randomly removing edges during training helps to reduce the effective receptive field of each node and disrupts excessive information diffusion, thus preserving local feature diversity.
    \item \textbf{Differential Operators}\cite{eliasof2021pde}: Smoothness-regularizing operators are applied to explicitly control the extent of feature mixing and preserve higher-order variations among nodes, especially in deep architectures.
\end{itemize}

While these techniques have demonstrated practical effectiveness, they often stem from empirical heuristics and lack a unified theoretical explanation. To address this, we introduce a new perspective grounded in condensed matter physics by drawing an analogy between over-smoothing and Anderson localization.This analogy provides a formal framework to understand how depth-induced homogenization in GNNs can be interpreted as a localization transition, and suggests new principles for designing models that resist over-smoothing.

\subsection{Anderson Localization in Condensed Matter Physics}
Anderson localization is a fundamental phenomenon in condensed matter physics that describes the localization of quantum wavefunctions in disordered systems.It was first proposed by Anderson in 1958 as a result of the interference effects between scattered waves in a disordered medium\cite{anderson1958absence,lagendijk2009fifty}.In a system with disorder, the wavefunctions of quantum particles, such as electrons, become spatially confined, preventing the free flow of particles and effectively "smoothing" the system's behavior.This phenomenon is particularly important in understanding the insulating behavior of materials at low temperatures.

The mathematical model for Anderson localization involves a Hamiltonian that includes both random on-site potentials and hopping terms between neighboring sites. The Hamiltonian for a system with disorder can be expressed as:

\begin{equation}
H = \sum_i \epsilon_i |i\rangle\langle i| + t \sum_{\langle i,j \rangle} (|i\rangle\langle j| + |j\rangle\langle i|)
\end{equation}

where $\epsilon_i$ is the random potential at site $i$, and $t$ is the hopping amplitude between neighboring sites.The random potentials introduce disorder into the system, which causes the wavefunction to localize at certain positions, depending on the strength of the disorder.As the disorder strength increases, the system undergoes a phase transition from a metallic phase, where wavefunctions spread across the system, to an insulating phase, where wavefunctions are confined to local regions.

The transition between these two phases is characterized by the disorder strength, which is often quantified by a dimensionless disorder parameter $\zeta$, representing the ratio of the disorder strength to the hopping amplitude.When $\zeta$ exceeds a critical value, the system transitions into the insulating phase, where the wavefunctions are localized. This transition is central to understanding many phenomena in condensed matter physics, such as the behavior of electrons in amorphous solids and high-resistance materials.

Recent extensions of Anderson's original theory have expanded the concept of localization to include non-Hermitian systems, systems with multiple disorder parameters, and systems in higher dimensions, further enhancing our understanding of how disorder leads to phase transitions in physical systems\cite{billy2008direct,nandkishore2015many,jiang2019interplay}.

\subsubsection*{Bridging the Gap: From Over-Smoothing to Localization}

The conceptual parallel between over-smoothing in GNNs and Anderson localization in disordered systems provides a unified physical interpretation of the homogenization effect in deep networks. In Anderson localization,increasing disorder suppresses high-frequency wave modes,leading to the spatial confinement of electron wavefunctions. Similarly,as the depth of a GNN increases,message passing disproportionately amplifies low-frequency spectral components while diminishing high-frequency signals, causing node features to converge and lose discriminative power. This process resembles a localization transition in spectral space.

By viewing the propagation of node features as analogous to the propagation of waves in a disordered medium, we can interpret over-smoothing as a form of spectral localization.This analogy enables us to import tools from condensed matter physics—such as participation ratio—to quantify and understand the spread (or confinement) of information in deep GNNs. Through this theoretical lens, we aim to not only explain the dynamics of over-smoothing but also motivate new architecture designs inspired by localization physics to preserve feature diversity across layers.

\begin{figure}[htbp]
    \centering
    \includegraphics[width=0.8\textwidth]{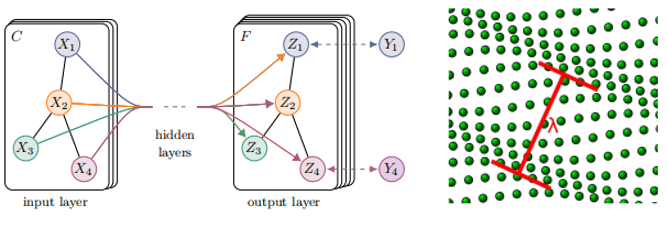}
    \caption{Message Passing in Graph Neural Networks and Wave Propagation in Disordered Systems\cite{kipf2016semi,wikipedia}}
    \label{fig:gnn_message_passing_wave_propagation}
\end{figure}

\section{Disordered Systems and  Graph Neural Networks}
In both physics and computer science, vibration modes in disordered systems and the spectral behavior of graph neural networks (GNNs) exhibit similarities in their structures and behaviors. The analysis of vibration modes in disordered systems is crucial for understanding the properties of materials, especially in irregular structures or materials like glass. In the spectral behavior of graph neural networks, nodes and edges represent the entities and their relationships in the graph, similar to how particles and their interactions are modeled in disordered systems. By drawing analogies between these two domains, we can better understand their deep connections and apply this understanding to the design and optimization of graph neural networks.

\begin{table}[h]
\centering
\caption{Dualities between disordered systems and graph neural networks.}
\begin{tabular}{@{}ll@{}}
\toprule
\textbf{Disordered System} & \textbf{Graph Neural Networks} \\ \midrule
Particles \( p_i \) & Nodes \( u_i \) \\
Particle Interactions \( F_{ij} \) & Edges \( \text{edge}_{ij} \) \\
Dynamical Matrix \( H \) & Laplacian Matrix \( L \) \\
Eigenvalues \( \omega^2 \) & Eigenvalues \( \lambda \) \\
Eigenvectors \( \mathbf{e}_i \) & Eigenvectors \( \mathbf{v}_i \) \\ \bottomrule
\end{tabular}
\end{table}

\setlength{\baselineskip}{13pt} 

\subsection{Particles and Nodes}
In spectral GNNs, nodes represent the entities in the graph, each associated with a feature vector \( \mathbf{X}_i \). Similarly, in disordered systems, particles are the fundamental units, each with a state represented by vectors \( \mathbf{u}_i \). The correspondence is given by:

\begin{equation}
\mathbf{X}_i \in \mathbb{R}^d \quad \text{(node feature vector)} \quad \longleftrightarrow \quad \mathbf{u}_i \in \mathbb{R}^d \quad \text{(particle state vector)}
\end{equation}
where \( d \) is the dimensionality of the vector.

\subsection{Particle Interactions and Edges}
Edges in a graph represent the relationships or interactions between nodes. These interactions can be encoded in the adjacency matrix \( A \), where \( A_{ij} = 1 \) if nodes \( i \) and \( j \) are connected, and \( 0 \) otherwise. In disordered systems, particle interactions are described by the interaction matrix \( F \), which captures the interaction strengths between particles. This correspondence is given by:

\begin{equation}
A = [A_{ij}], \quad A_{ij} = \begin{cases} 
1, & \text{if nodes } i \text{ and } j \text{ are connected} \\
0, & \text{otherwise}
\end{cases}
\end{equation}

\begin{equation}
F = [F_{ij}], \quad F_{ij} = \text{interaction strength between particle } i \text{ and } j
\end{equation}

Thus, edges in the graph correspond to particle interactions in disordered systems.

\subsection{Laplacian Matrix and Dynamical Matrix}
The  normalized Laplacian matrix \( L \) in spectral GNNs is defined as:
\begin{equation}
L = D^{-1/2} (D - A) D^{-1/2} = I - D^{-1/2} A D^{-1/2}
\end{equation}
where \( D \) is the degree matrix, and \( A \) is the adjacency matrix. The Laplacian matrix captures the connectivity structure of the graph.
In disordered systems, the dynamical matrix \( H \) describes the interactions between particles and governs the system's vibrational behavior. It is derived from the second derivative of the potential energy \( \Phi \):
\begin{equation}
H = \left[ \frac{\partial^2 \Phi}{\partial R_l \partial R_{l'}} \right]
\end{equation}
where \( \Phi \) is the potential energy, and \( R_l \) represents the position of particle \( l \). The corresponding equation for the vibration modes is:
\begin{equation}
H \mathbf{e}_i = \omega^2 \mathbf{e}_i
\end{equation}
In spectral GNNs, the eigenvalues \( \lambda \) of the Laplacian matrix \( L \) represent the frequencies of the graph’s signal, and the eigenvectors \( \mathbf{v}_i \) represent the signal components at those frequencies. Similarly, in disordered systems, the eigenvalues \( \omega^2 \) of the dynamical matrix \( H \) correspond to the vibration frequencies, and the eigenvectors \( \mathbf{e}_i \) represent the vibration modes.

For the Laplacian matrix, the eigenvalue problem is:
\begin{equation}
L \mathbf{v} = \lambda \mathbf{v}
\end{equation}
For the dynamical matrix, the vibration mode equation is:
\begin{equation}
H \mathbf{e}_i = \omega^2 \mathbf{e}_i
\end{equation}
Thus, the eigenvalues \( \lambda \) of the Laplacian matrix correspond to the vibration frequencies \( \omega \) in disordered systems, and the eigenvectors \( \mathbf{v} \) correspond to the vibration modes \( \mathbf{e}_i \).

\section{Understanding Over-Smoothing in Graph Neural Networks Based on Anderson Localization}

\subsection{Over-Smoothing Accompanied by the Decrease in Participation Degree of High-Frequency Graph Signals}

In disordered systems, the participation degree (\( p_{\omega} \)) is a measure used to quantify the spatial extent of vibration modes. It is defined as the normalized sum of the squared components of the vibration mode \( \mathbf{e}_i \) across all particles \( i \), which characterizes how widely a vibration mode is distributed across the system. The participation degree is crucial for understanding the nature of localization in disordered systems, such as Anderson localization. The participation degree \( p_{\omega} \) for the \( \omega \)-th mode is mathematically expressed as:

\begin{equation}
p_{\omega} = \frac{\left( \sum_i e_{\omega,i}^2 \right)^2}{N \sum_i e_{\omega,i}^4}
\end{equation}

where \( e_{\omega,i} \) is the \( i \)-th component of the vibration mode at frequency \( \omega \), and \( N \) is the total number of particles. For a perfectly delocalized mode, \( p_{\omega} \approx 1 \), while for a localized mode, \( p_{\omega} \to 0 \), as the mode becomes confined to a small region of the system. In Anderson localization, low-frequency modes are often partially localized (i.e., \( p_{\omega} \) is small but nonzero), while high-frequency modes exhibit complete localization (i.e., \( p_{\omega} \to 0 \)).

Thus, the participation degree provides a crucial metric for understanding the transition from delocalized to localized states in disordered systems, with its behavior being a hallmark of localization phenomena.

\begin{figure}[htbp]
    \centering
    \includegraphics[width=0.8\textwidth]{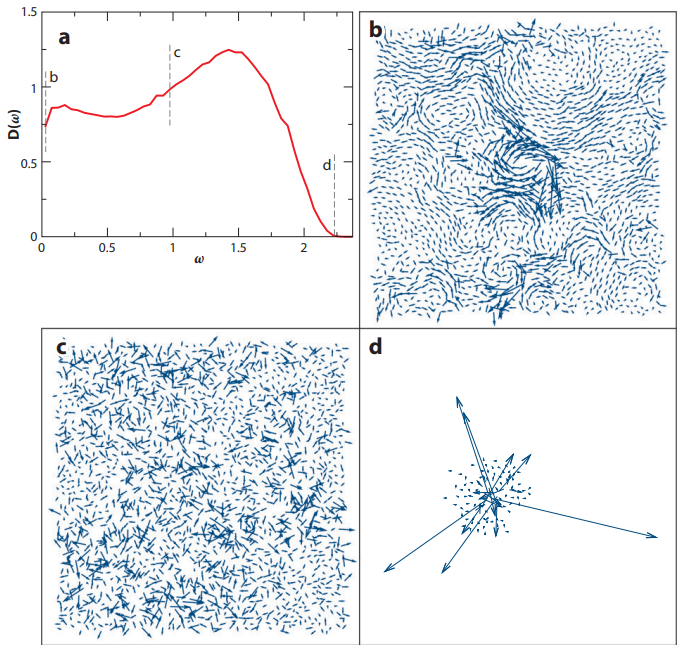} 
    \caption{\small Vibration modes in a two-dimensional disordered harmonic interaction system with 2000 particles and a distance point J of body integral \( \Delta \phi = 10^{-4} \). (a) The density of states \( D(\omega) \), with vertical lines marking the frequencies of the vibration modes shown in (b)-(d). The spatial distribution of the vibration modes: (b) low-frequency quasi-localized mode, (c) medium-frequency anomalous extended mode, and (d) high-frequency localized mode. The arrows in the figure represent the component of the vibration mode on a particular particle \cite{liu2010jamming}.}
    \label{fig:vibration_modes}
\end{figure}

In the context of Graph Neural Networks (GNNs), we draw an analogy between participation degree in disordered systems and the phenomenon of over-smoothing. In GNNs, over-smoothing refers to the situation where node features become indistinguishable after passing through multiple layers of graph convolutions. This happens when the node features are averaged too much, resulting in a loss of discriminative power between nodes. This homogenization effect in GNNs is similar to the localization phenomenon observed in disordered systems.

To define a participation degree in GNNs, we consider the eigenvectors of the Laplacian matrix \( L \) in spectral GNNs, as they describe the signal propagation across the graph. The Laplacian matrix \( L \) can be diagonalized as:

\begin{equation}
L = V \Lambda V^T
\end{equation}

where \( V = [v_1, v_2, \dots, v_N] \) is the matrix of eigenvectors of the Laplacian matrix \( L \), and \( \Lambda = \text{diag}(\lambda_1, \lambda_2, \dots, \lambda_N) \) is the diagonal matrix of eigenvalues corresponding to the eigenvectors. The eigenvectors \( v_i \) are orthonormal, and the eigenvalues \( \lambda_i \) correspond to the frequencies of the graph's signals.

Let \( \mathbf{h}_i \) be the feature vector of node \( i \). To project the feature vector of node \( i \) onto the eigenvector \( v_{\lambda} \) corresponding to the eigenvalue \( \lambda \), we compute the dot product between the node feature \( \mathbf{h}_i \) and the eigenvector \( v_{\lambda} \) corresponding to the frequency component \( \lambda \). The projection of the node feature \( \mathbf{h}_i \) onto the eigenvector \( v_{\lambda} \) is given by:

\begin{equation}
h_{\lambda,i} = \mathbf{h}_i^T v_{\lambda}
\end{equation}

This represents the component of the node feature \( \mathbf{h}_i \) along the eigenvector corresponding to the frequency \( \lambda \). The participation degree \( p_{\lambda} \) can then be defined as follows, using the projection of the node features onto the eigenvectors:

\begin{equation}
p_{\lambda} = \frac{\left( \sum_i h_{\lambda,i}^2 \right)^2}{N \sum_i h_{\lambda,i}^4}
\end{equation}

where \( N \) is the number of nodes in the graph, and \( h_{\lambda,i} \) is the projection of the feature vector of node \( i \) onto the eigenvector \( v_{\lambda} \). This formula captures how much the feature component corresponding to the frequency \( \lambda \) is spread across all the nodes in the graph.

- A low participation degree \( p_{\lambda} \to 0 \) indicates that the feature is highly localized to a small subset of nodes, implying that the signal is dominated by low-frequency components (similar to localized modes in Anderson localization).
- A high participation degree \( p_{\lambda} \approx 1 \) suggests that the signal is more evenly distributed across the entire graph, implying that the signal is more global (similar to delocalized modes in Anderson localization).

Thus, \( h_{\lambda,i} \) is obtained through the projection of the node features onto the eigenvectors of the Laplacian matrix, capturing the frequency components of the graph’s signal. This participation degree measure allows us to analyze the propagation and localization of information in GNNs.

\begin{figure}[htbp]
    \centering
    \includegraphics[width=0.8\textwidth]{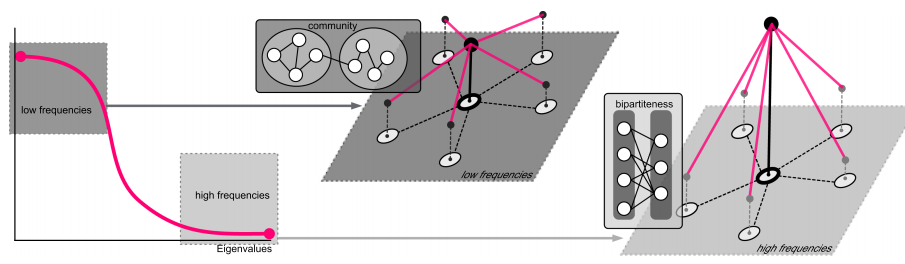}
    \caption{Spectral Decomposition in Graph Neural Networks \cite{chen2023bridging}.}
    \label{fig:Spectral Decomposition}
\end{figure}

In GNNs, over-smoothing occurs when node features become homogenized after multiple layers of message passing, leading to a situation where all nodes in the graph have indistinguishable features. This corresponds to the loss of distinctive features as the node representations converge, much like how in disordered systems, the low-frequency components of the vibration modes spread across the entire system while high-frequency components become suppressed. As the depth of the GNN increases, each layer aggregates features from neighboring nodes, diffusing information across the graph. However, if the participation degree of the features at higher layers becomes very low (\( p_{\lambda} \to 0 \)), it indicates that the features are being dominated by low-frequency components, leading to a loss of node individuality. This behavior can be formally expressed as:

\begin{equation}
\lim_{L \to \infty} p_{\lambda} \to 0 \quad \forall \lambda \in [0, 2]
\end{equation}

where \( L \) denotes the number of layers. As the network depth increases, the participation degree decreases, meaning the signal is increasingly dominated by low-frequency components, causing the features of the nodes to become more similar to each other, leading to over-smoothing. This effect mirrors the way high-frequency components become suppressed in disordered systems, while low-frequency components dominate.In summary, the participation degree in GNNs can serve as a metric for understanding over-smoothing, with a low participation degree indicating that the node features are being dominated by low-frequency components. This contributes to the homogenization of node representations. By carefully managing the participation degree in GNNs, it is possible to mitigate over-smoothing and maintain distinctive node representations.

\subsection{The high-frequency graph signals need to overcome higher potential barriers to overcome over-smoothing.}

As shown in Figure 4, in the low-frequency region, the participation degree is small but not zero, corresponding to quasi-localized vibration modes. In the mid-frequency region, the participation degree is larger, corresponding to extended anomalous vibration modes. In the high-frequency region, the participation degree tends to zero, corresponding to Anderson localization. At the same time, the vibration modes in the high-frequency region face the highest energy potential barriers, which can be seen as the cause of Anderson localization. Accordingly, as a reasonable inference, we believe that the over-smoothing in graph neural networks is caused by the high-frequency components of the graph signals being unable to overcome certain potential barriers, leading to limited propagation.

\begin{figure}[htbp]
    \centering
    \includegraphics[width=0.8\textwidth]{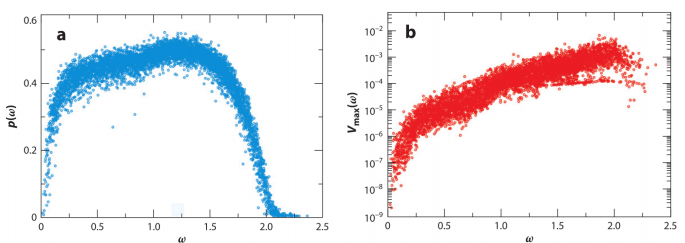}
    \caption{Three-Dimensional Bimodal Harmonic Interaction System Configuration with N = 1000 Particles and Body Integral \( \Delta \phi = 0.1 \) at a Distance Point J: (a) Vibration Mode Participation Degree \( p(\omega) \), (b) Energy Potential Barriers \( V_{\text{max}}(\omega) \) Encountered When Entering Another Energy Well Along the Vibration Mode Direction. It can be seen that low-frequency vibration modes with small participation degrees face the lowest energy barriers before recombination\cite{liu2010jamming}.}
    \label{fig:Potential Barriers}
\end{figure}

\subsection{The localization length limits the range of graph signal propagation.}

In Anderson localization, disorder plays a crucial role in determining the localization length, which is the length scale over which a quantum wavefunction remains extended before it becomes localized. The relationship between disorder strength and localization length is fundamental to understanding the behavior of systems subject to random potentials. Specifically, the localization length \( \xi \) decreases as the disorder strength \( \Delta \epsilon \) increases. This can be expressed mathematically as:

\begin{equation}
\xi \sim (\Delta \epsilon)^{-\gamma}
\end{equation}

where \( \gamma \) is a constant that depends on the system's dimension and the specifics of the disorder. In low-dimensional systems (e.g., 1D or 2D), the localization length is particularly sensitive to disorder, meaning that even small amounts of disorder can cause the wavefunction to become localized over short distances. In contrast, in high-dimensional systems, larger amounts of disorder are generally required for the system to undergo localization.

The localization length is a measure of how far the wavefunction can extend before becoming localized. As disorder increases, the localization length shrinks, and the system tends toward a localized phase. This phenomenon can be understood as the tendency of wavefunctions to become increasingly confined in space due to the random potential landscape introduced by disorder.

Now, applying this idea to graph signals, we hypothesize that the disorder in a graph affects the high-frequency components of the graph signal in a similar manner to how disorder affects wavefunctions in Anderson localization. In a graph neural network, the high-frequency components of the graph signal correspond to modes that change rapidly across the graph, analogous to high-frequency vibration modes in a disordered physical system. These high-frequency components are more susceptible to "localization" effects, where their influence is confined to a small region of the graph.

We conjecture that, just as in Anderson localization, the disorder in the graph—represented by the irregularity in the graph structure or the edge weights—causes the high-frequency components of the graph signal to become increasingly localized. This leads to an over-smoothing effect, where the signal fails to propagate effectively across the entire graph, as the localized components do not contribute to the overall message passing beyond a small neighborhood.

Thus, as the disorder in the graph increases, the localization length of the graph signal decreases. This implies that the message passing in graph neural networks becomes more limited, with high-frequency components being increasingly "trapped" in local regions of the graph. Consequently, the higher the disorder in the graph, the more significant the Anderson-like localization behavior, and the more pronounced the over-smoothing effect becomes.

This hypothesis suggests that the over-smoothing issue in graph neural networks may be a result of the inability of high-frequency components of the graph signal to "overcome" certain barriers in the graph's structure, much like how high-frequency vibration modes in a disordered system face higher energy barriers leading to localization.

\subsection{The Order Parameter of Graph Neural Networks Inspired by Disordered Systems.}

In disordered systems, the degree of disorder is often characterized by the spatial fluctuations in the coordination number, which refers to the variation in the number of neighboring particles (or sites) that interact with a given particle (or site). This fluctuation is crucial in determining the behavior of the system, especially in relation to the localization transition. Similarly, for graph neural networks (GNNs), we can define a measure of disorder in the graph structure, which will influence the behavior of graph signal propagation and potentially lead to over-smoothing in high-frequency components.

\textbf{Spatial Domain Disorder}

Inspired by the concept of coordination number fluctuations in disordered systems, we define the \textbf{degree fluctuation} \( \Delta k \) of a graph as a measure of the spatial disorder. For each node \( i \) in the graph, the degree \( k_i \) represents the number of edges connecting to it, and the degree fluctuation is given by the standard deviation of the node degrees:

\begin{equation}
\Delta k = \frac{1}{N} \sqrt{ \sum_{i=1}^{N} \left( k_i - \langle k \rangle \right)^2 }
\end{equation}

where \( \langle k \rangle \) is the average degree of the graph, and \( N \) is the total number of nodes. The degree fluctuation \( \Delta k \) measures the variation in node connectivity, analogous to spatial fluctuations in disordered systems. Higher degree fluctuations imply higher disorder in the graph, potentially leading to more localized graph signals, especially in higher-frequency components, similar to how disorder in physical systems leads to localization of wavefunctions.

\textbf{Hypothesis: Disorder and Over-Smoothing}

Drawing from the analogy with disordered physical systems, we conjecture that the graph's disorder, as quantified by spatial fluctuations, limits the range of graph signal propagation. Specifically, as the disorder in the graph increases—measured by degree fluctuations—the localization length of the graph signal decreases, leading to over-smoothing in graph neural networks. The high-frequency components of the graph signal become increasingly localized, which corresponds to the inability of these components to propagate effectively across the entire graph, thus resulting in over-smoothing.

This definition of disorder in the spatial domain provides a framework for understanding how the graph structure impacts signal propagation, and how high disorder in the graph can lead to limited message passing, analogous to Anderson localization in physical systems. The higher the disorder in the graph, the more pronounced the localization behavior becomes, and the more likely over-smoothing will occur in graph neural networks.

\subsection{Reducing disorder alleviates over-smoothing.}

Inspired by Anderson localization, the localization length is related to the degree of disorder in the system. As the disorder increases, the likelihood of Anderson localization behavior occurring increases. In the context of graph neural networks, we hypothesize that over-smoothing can be alleviated by reducing the disorder in the information flow during the aggregation process. Specifically, we aim to introduce a mechanism that modulates the flow of information based on the degree of disorder in the graph.

For example, for nodes with high degrees \( k_i \), there is a certain probability \( P_{\text{no-agg}}(k_i) \) that each edge will be disconnected during the information aggregation process, meaning that the node will not aggregate information from some of its neighbors. This probability can be expressed as:

\begin{equation}
P_{\text{no-agg}}(k_i) = \frac{1}{1 + e^{-\alpha(k_i - \langle k \rangle)}}
\end{equation}

where \( \alpha \) is a parameter that controls the probability decay, and \( \langle k \rangle \) is the average degree of the graph. This function ensures that nodes with higher degrees are more likely to disconnect edges and not aggregate all the information from their neighbors.

On the other hand, for nodes with lower degrees \( k_i \), there is a certain probability \( P_{\text{agg}}(k_i) \) that they will include information from nodes that are not directly connected to them, effectively adding non-neighboring nodes to the aggregation process. This probability can be modeled as:

\begin{equation}
P_{\text{agg}}(k_i) = \frac{1}{1 + e^{\alpha(k_i - \langle k \rangle)}}
\end{equation}

Here, nodes with lower degrees have an increased probability of adding non-neighboring nodes' information, helping to diversify the information flow and prevent over-smoothing.

This approach essentially reduces the disorder in the information aggregation process by lowering the degree fluctuation in the graph. The degree fluctuation, which can be quantified by the standard deviation of node degrees, measures how much the degree of each node deviates from the average degree of the graph. By controlling the aggregation process to reduce this fluctuation, we can lower the disorder in the graph, which in turn reduces the likelihood of over-smoothing.

This is a potential hypothesis, and in the future, we plan to validate this approach through experiments to confirm the effectiveness of reducing disorder as a strategy for alleviating over-smoothing in graph neural networks.

By reducing the disorder in the information aggregation process, we can effectively reduce the likelihood of over-smoothing, ensuring that high-frequency components of the graph signal can propagate more effectively across the network, thereby alleviating the over-smoothing issue.

\section{Conclusion}

In this paper, we have proposed a novel approach to understanding over-smoothing in Graph Neural Networks (GNNs) by drawing an analogy to Anderson localization in disordered systems. By introducing the concept of the \textbf{participation degree}, we have developed a framework to quantify the phenomenon of over-smoothing and understand its underlying mechanisms. We showed that, similar to the localization phenomenon in disordered systems, the increasing depth of GNNs leads to the homogenization of node features, where low-frequency components dominate and high-frequency components become suppressed. This results in a loss of discriminative power among nodes, akin to the transition from a metallic phase to an insulating phase in disordered systems.
Our analysis suggests that over-smoothing in GNNs can be viewed as a form of spectral localization, where the high-frequency graph signals face localization barriers, limiting their propagation across the graph. This connection to Anderson localization provides a fresh perspective on managing over-smoothing in GNNs.Currently, the research presented in this paper remains at the theoretical analysis level. Looking ahead, a key direction for future research is to validate our hypothesis through experiments and expand the concept of participation degree into a broader set of order parameters. These parameters could be used to analyze the convergence and generalization of GNNs more rigorously, as well as guide the design of models that are robust to over-smoothing. Additionally, applying this framework to different types of GNN architectures and datasets will allow for more targeted optimizations and better adaptation to various graph-structured data.Based on the insights drawn from Anderson localization, we aim to design effective mechanisms or new GNN architectures that can alleviate over-smoothing, ultimately enhancing the efficiency, scalability, and performance of GNNs in real-world applications.

\bibliographystyle{unsrt}
\bibliography{ref}
\end{document}